**Causal-Privacy Audit Workflow for Synthetic and Distilled Data in Dropout Support**

Hanghang Zheng[1$], Xiwei Zhuang[2], Zhong Wang[2], Hong Liu[2], Xiao Chen[2],

Jingwen He[2], Xia Li[2,3]

---

[1] Central University of Finance and Economics, Beijing, PR China
[2] Central University of Finance and Economics, Beijing, PR China
[3] China Development Bank, Beijing, PR China
[4] University of Cambridge, Cambridge, CB2 3EB United Kingdom

## Abstract

Synthetic and distilled student data are increasingly used to enable privacy-conscious learning analytics, yet their suitability for decision-facing institutional support remains uncertain. In dropout support, generated data must preserve not only predictive utility or distributional resemblance, but also the financial-status evidence used to guide advising, payment-plan assistance, and scholarship-related decisions. Method: This study introduces CaP-Eval, a decision-facing causal-privacy audit workflow for evaluating generated student data under a fixed estimand, timing-aware adjustment design, estimator set, and empirical privacy-governance screen. The workflow compares original, distilled, adversarial synthetic, statistical synthetic, and DPGNet privacy-oriented generated data on predictive utility, treatment-effect fidelity, robustness to alternative estimators, and local training-record proximity. Results: DPGNet and distilled data preserved the original financial-status treatment-effect structure more reliably than the adversarial and Gaussian Copula baselines. DPGNet preserved full direction and rank agreement across epsilon levels; epsilon = 10 produced the smallest non-original IPW and DML deviations, while epsilon = 1 and epsilon = 5 amplified several financial-status contrasts. Distilled data remained highly faithful but retained the strongest local training-record proximity signal. TabularGNet preserved qualitative directions with moderate attenuation, and Gaussian Copula compressed effect magnitudes. Conclusions: Predictive utility, privacy orientation, empirical disclosure signals, and causal fidelity diverged; generated student data require joint audits of direction, magnitude, overlap, and release-governance risk before decision use.

## 1. Introduction

Student dropout remains a persistent challenge in higher education because it affects student attainment, institutional performance, and the allocation of advising and financial-support resources [1]. Prior learning-analytics studies show that dropout risk can be predicted from demographic, socioeconomic, academic, behavioral, and learning-management-system variables[2],[3],[4],[5],[6],[7]. These models can support early-warning systems and institutional monitoring, particularly when their outputs are interpretable to advisors and decision makers[8],[9],[10]. However, predictive accuracy alone does not indicate which institutional conditions should become targets for action. A variable may be predictive because it is measured late, proxies disengagement, or captures background differences, even when it is not an appropriate support target[11],[12],[13]. Retention analytics therefore needs to distinguish prediction from decision-relevant evidence.

This distinction becomes more important when analysis relies on synthetic, distilled, or differentially private data rather than original student records. Such data representations can reduce direct exposure of sensitive administrative information and support reproducible or collaborative analysis, but they can also alter the treatment-covariate-outcome relationships that guide institutional decisions. We therefore frame the problem as a decision-facing audit rather than as a utility-only benchmark: can a generated dataset support the same dropout-support conclusions that would be drawn from the original training data under a fixed estimand, timing rule, covariate set, and estimator family? We term this audit workflow CaP-Eval (Causal-Privacy Evaluation).

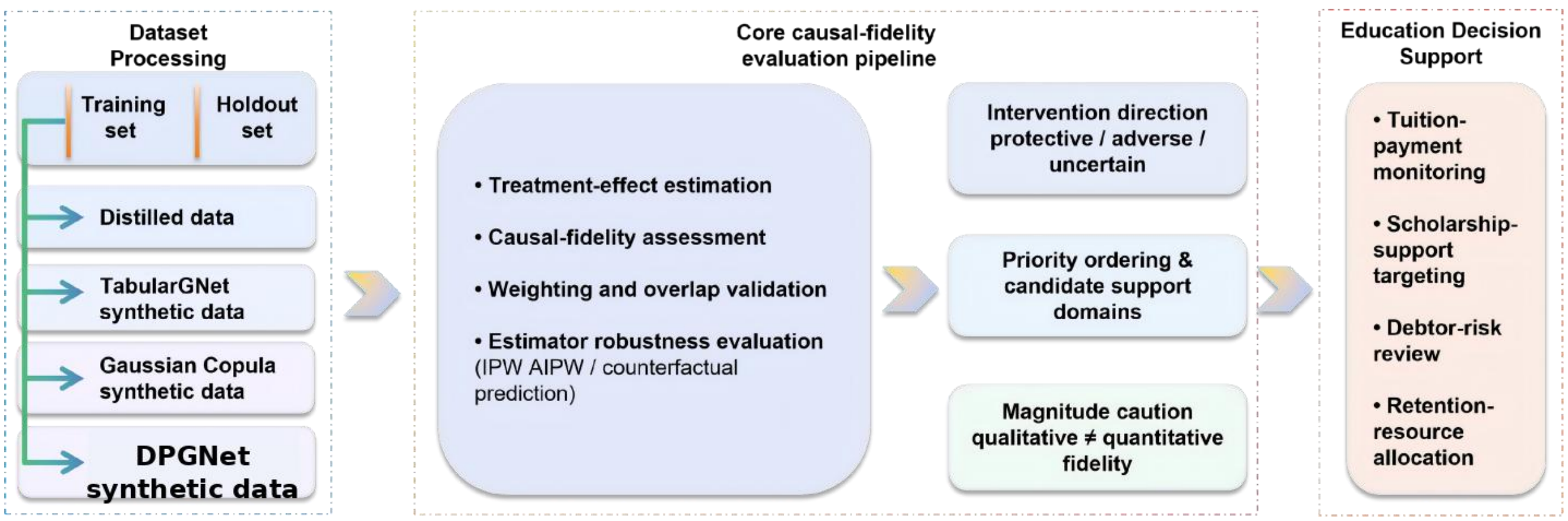

Figure 1. Workflow of this study. Original, distilled, adversarial, Gaussian Copula, and DPGNet datasets are compared through timing-aware treatment-effect estimation, overlap and weighting evaluation, causal-robustness assessment, estimator checks, and privacy review. Causal robustness and empirical privacy-governance evidence are reported together.

We compare the synthetic and distilled datasets under a common adjustment design so that differences can be attributed primarily to the data-generation process[14]. In observational studies, estimated effects depend on adjustment choices, overlap, and evaluation criteria. Holding these elements fixed allows differences in conclusions to be attributed to the datasets themselves rather than to changes in the analysis pipeline.

We examine this problem using the Students' Dropout and Academic Success dataset, which combines demographic, socioeconomic, enrollment, macroeconomic, financial, and semester-level academic variables[15]. Its richness is useful, but it also creates a timing problem that is easy to miss in predictive work. Semester-performance variables are observed after enrollment and may partly respond to financial conditions; controlling for them would change the question from whether financial status is associated with dropout risk to whether it matters after later academic performance has already unfolded[12]. The main adjustment set therefore uses enrollment-time covariates and excludes later curricular-unit measures. The same rule is applied to every generated or distilled dataset.

The analysis focuses on scholarship-holder status, tuition-fee payment status, and debtor status. These variables are not abstract risk markers. They correspond to concrete institutional processes: financial-aid review, payment-plan support, debt-related advising, and outreach when payment status changes. This is also why education differs from many benchmark studies of synthetic data. A small distortion in a financial-status association can change which students are flagged for support, how scarce advising time is allocated, or whether a payment-related signal is treated as a need for assistance rather than as a late-stage correlate of dropout.

The privacy question is evaluated in parallel. Synthetic and distilled datasets are often introduced to reduce exposure of sensitive student records, but privacy reassurance based only on the absence of exact duplicates is too weak for this setting. We therefore check exact record reproduction, nearest-record distances, train-holdout distance dominance, membership-inference proxies, quasi-identifier exposure, and attribute-disclosure advantage. Distilled data receive particular scrutiny. Dataset condensation may reduce direct raw-record release[16], yet later work shows that local or worst-case privacy risks can remain visible even when average evaluations look benign[17],[18]. We treat these evaluations as evidence for disclosure review, not as a replacement for formal privacy guarantees.

To our knowledge, few studies have jointly evaluated causal fidelity and empirical privacy risk for generated student data across distillation, adversarial synthesis, statistical synthesis, and DPGNet privacy-oriented generation within a unified formal higher-education decision-support workflow. Existing learning-analytics work has largely emphasized predictive utility, distributional resemblance, or privacy screening, whereas causal synthetic-data studies in other domains do not directly resolve the tuition-bearing, credential-linked, and institutionally accountable nature of formal higher-education dropout decisions. This gap is consequential: a dataset that reproduces classifier accuracy may still distort the evidence an institution uses to decide where to direct financial-support resources.

The paper makes four contributions. First, it formalizes CaP-Eval as a decision-facing audit workflow for asking whether generated student data preserve the financial-status evidence institutions use to direct support. Second, it separates directional validity from effect-size fidelity through direction preservation, rank fidelity, mean absolute effect deviation, and worst-case treatment gap. Third, it shows that prediction, distributional resemblance, privacy orientation, empirical disclosure evidence, and treatment-effect magnitude can diverge in formal higher-education dropout data. Fourth, it translates these differences into tiered release-and-use guidance for internal validation, workflow testing, external development, and student-facing decision support.

## 2. Related work

Student dropout has been widely studied in educational data mining and learning analytics[19]. Prior work shows that demographic, socioeconomic, enrollment, academic, behavioral, and learning-management-system variables can support early-warning models[1],[4],[6],[20],[21],[22]. Recent studies have further compared logistic regression, support vector machines, random forests, gradient boosting, and neural-network-based models for dropout prediction[7],[23],[24],[25]. However, prediction and intervention remain distinct. A variable may improve classification without representing a feasible or desirable support target. This distinction is especially important in on-campus higher-education management, where early-warning outputs are expected to inform advising, academic support, financial-aid review, payment-plan assistance, or early outreach[8]. Learning analytics is intended to generate actionable insights that improve learning and teaching rather than stop at descriptive or predictive monitoring[26],[27],[28]. Dropout analytics should therefore evaluate whether the variables and conclusions used by a model are relevant to practical educational-management decisions.

### 2.1 Synthetic data for education

Synthetic data offer one response to restricted access to educational records[29],[30],[31],[32]. In learning analytics, synthetic tabular data are commonly evaluated through resemblance to the original data, downstream utility, and privacy risk[29],[32],[33],[34],[35]. The present study extends this line of evaluation by asking a decision-facing question that utility/privacy benchmarks alone do not answer: whether generated or distilled data preserve the financial-status associations that would guide institutional dropout-support decisions. Differentially private generative models, including gradient-perturbed adversarial generators and autoencoder/variational generative models, seek to reduce individual-record exposure while retaining useful synthetic-data structure[36],[37]. The DPGNet condition evaluated here is therefore treated as a privacy-oriented generated-data stress test, not as a universal benchmark of all differentially private synthesizers. Recent causal-synthetic-data work further shows that privacy and prediction criteria are insufficient for causal or intervention-oriented use: a generator can resemble the source data while distorting average treatment effects[38]. This motivates a causal-fidelity audit rather than a fidelity-only or privacy-only evaluation.

### 2.2 Dataset distillation and privacy

Dataset distillation is related to synthetic data but has a different privacy and utility profile. It compresses a dataset into a smaller support set while retaining learning-relevant structure[39]. Early work argued that dataset condensation may help privacy by replacing raw examples with optimized synthetic supports[16]. Later analyses warned that such claims require careful scrutiny because empirical privacy evaluations can miss vulnerable samples or local leakage[17],[18]. Recent distillation theory further shows that soft labels and teacher-student transfer can leak memorized knowledge even when the student never directly observes the held-out examples[40]. These findings suggest that strong analytic fidelity in distilled data may be accompanied by stronger local training-data retention, which is precisely the trade-off evaluated in this study.

### 2.3 Causal inference in educational analytics

Causal inference is increasingly relevant to educational analytics because institutions need to know not only which students are at risk, but also which conditions may be worth targeting[10].

Prior studies have applied propensity-score weighting, matching, graphical assumptions, and machine-learning-based causal methods to educational outcomes[13],[41],[42],[43],[44]. These studies also show that causal estimates are sensitive to timing and adjustment choices[12]. In dropout analysis, this issue is especially important because post-enrollment academic-performance variables can be highly predictive while also changing the interpretation of financial-status estimates when used as controls.

### 2.4 Formal higher-education dropout versus MOOC non-completion

MOOC datasets and online learning logs are valuable for studying engagement, persistence, platform behavior, and large-scale learning trajectories[45],[46]. However, they are not direct substitutes for formal higher-education dropout data. In MOOCs, dropout may refer to non-starting, partial participation, non-certification, or leaving after meeting a personal learning goal[46]. In formal degree programs, dropout is tied to institutional enrollment, academic progression, tuition payment, credential attainment, and student-support decisions[5],[6]. These differences change both the outcome meaning and the estimand; therefore, this study focuses on formal higher education.

## 3. Methodology

### 3.1. Data preparation

The analysis used the Students' Dropout and Academic Success dataset, which contains 4,424 formal higher-education student records covering demographic, socioeconomic, enrollment, macroeconomic, financial-status, academic-pathway, and outcome variables[15]. The dataset is useful for a decision-facing audit because it combines mixed continuous, binary, and categorical administrative variables; an imbalanced binary dropout outcome after recoding; financial-status indicators that are plausible institutional action signals; and post-enrollment academic variables that are highly predictive but potentially downstream of financial status. These features make it a compact but realistic stress case for formal higher-education administrative data, where prediction, timing, confounding, and release governance interact.

Because the dataset includes variables measured at different stages of the student trajectory, the adjustment set was defined by measurement timing rather than by predictive importance. Baseline covariates included demographic, socioeconomic, enrollment-path, parental-background, and macroeconomic variables, such as application mode, course, previous qualification, parental qualification and occupation, gender, age at enrollment, unemployment rate, inflation rate, and GDP. First- and second-semester curricular-unit variables were excluded because they are post-enrollment measures and may lie downstream of financial status[12]. This timing rule was applied consistently across the original, distilled, adversarial, Gaussian Copula, and DPGNet regimes. Figure 2 shows the timing-aware adjustment design used in the main analysis.

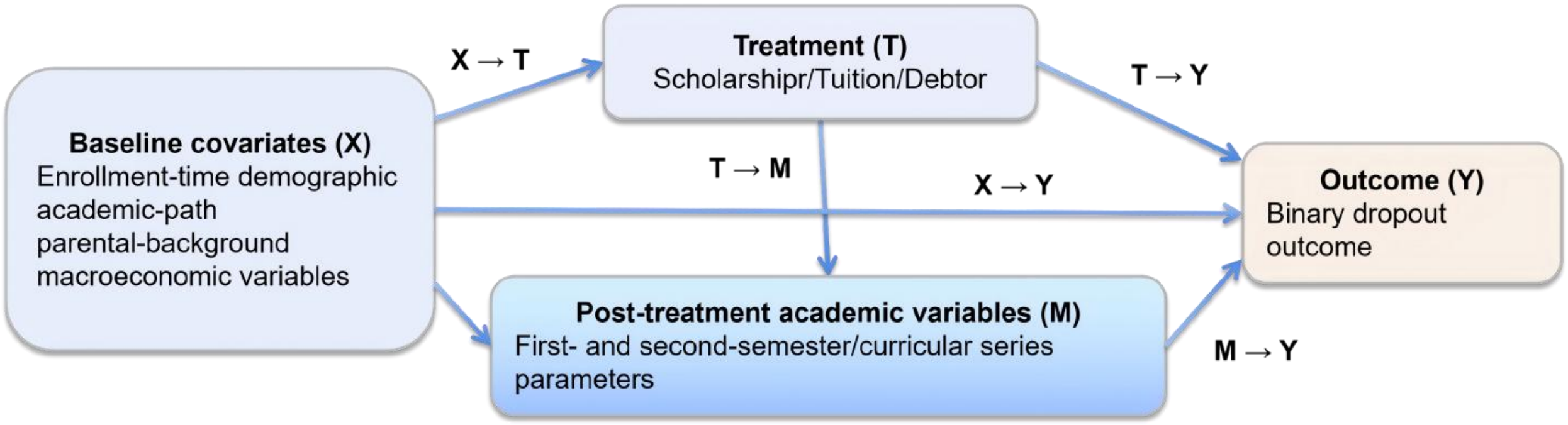

**Figure 2. Timing-aware directed acyclic graph for the observational adjustment design.** Baseline covariates are eligible for adjustment. The treatment represents one of the three financial-status indicators: scholarship holder status, up-to-date tuition status, or debtor status. First- and second-semester academic-performance variables are treated as later academic descendants and are excluded from the main adjustment set. The outcome is binary dropout status.

The original dataset was divided into a stratified training set and holdout set using an 80/20 split with a pre-specified seed of 42. The training set was used to construct the distilled and synthetic datasets, while the holdout set was kept separate from the dataset generation process. The training set served as the original reference regime for the treatment-effect and robustness analyses.

Six non-original data regimes were generated from the training set only; the holdout set was never used for distillation, synthetic-data fitting, or DPGNet generation. The distilled dataset was produced using a KIP-style kernel-based tabular distillation procedure[39] with a main support size of 1,000 and a support-size ablation grid of 10, 50, 100, 300, 500, 750, and 1,000 support points. The procedure used standardized numeric features, learnable support labels, class-balanced support initialization, and post-processing of integer and binary columns back to the original schema. Additional implementation settings are provided in the supplementary materials.

The adversarial synthetic dataset was generated using an SDV CTGAN-based TabularGNet baseline[47],[48] with automatically detected single-table metadata, 200 training epochs, fixed random state 42, enforce_rounding = False, and CPU-preferred deterministic execution when supported. The Gaussian Copula baseline used the same training pool and SDV metadata workflow[47], with enforce_min_max_values = True, enforce_rounding = False, and sample sizes matched to the original training regime. For both TabularGNet and Gaussian Copula, the main synthetic sample size matched the original training-set size, while additional sample budgets were retained only for supplementary sensitivity checks.

The DPGNet generator was evaluated at epsilon = 1, 5, and 10 with delta = 1 × 10^-5, fixed random state 42, and the same training-only data boundary used for the other non-original regimes. The evaluated DPGNet configuration was an epsilon-indexed differentially private generator rather than a teacher-aggregation generator; teacher-model counts and teacher epochs were therefore not applicable. The privacy budget was the planned varying factor across DPGNet conditions, while preprocessing, output schema, sample size, and evaluation pipeline were held fixed. Post-generation schema restoration followed the original mixed-type table: numeric variables were returned to the original scale, integer and binary fields were rounded and cast back to legal values, and categorical outputs were restricted to observed category support. Because optimizer-level quantities such as clipping norms, noise schedules, and accountant internals were not exposed, they were not inferred from generated outputs. We therefore report DPGNet as an implementation-specific privacy-oriented stress test rather than as a universal ranking of differentially private synthesizers[36],[37].

### 3.2. Predictive and treatment-effect evaluation

Before estimating financial-status associations, we specified a fixed timing-aware design for all treatments. Scholarship-holder status, tuition-fee payment status, and debtor status were evaluated against the same binary dropout outcome. Baseline enrollment-time covariates were used for adjustment, while first- and second-semester curricular-unit variables were excluded because they may occur downstream of the financial-status indicators. For tuition-fee payment and debtor status, a sensitivity specification additionally adjusted for scholarship-holder status. This design aligns the analysis with decision-support use rather than with pure predictive optimization.

Crucially, the generation and causal-audit stages were deliberately decoupled. The synthetic and distilled datasets were generated from the full training table so that each regime could preserve the high-dimensional joint data geometry available to a realistic generation process. During the treatment-effect audit, however, the conditioning sets were restricted to the pre-treatment baseline covariates defined in the timing-aware DAG, while post-enrollment academic-performance variables were excluded. This design separates generative fidelity from causal identification and reduces the risk of overcontrol or mediator contamination when evaluating the financial-status estimand.

Predictive performance was evaluated using AUC, average precision, precision, recall, F1-score, and Brier score. Higher AUC, average precision, precision, recall, and F1-score indicate stronger predictive performance, while a lower Brier score indicates more accurate predicted probabilities.

Financial-status associations were estimated separately for scholarship-holder status, tuition-fee payment status, and debtor status. Stabilized inverse probability weighting was used as the main estimator. Propensity scores were estimated using logistic regression with the timing-aware baseline covariates, class balancing, and a maximum of 2,000 iterations. To reduce the influence of extreme weights and improve weight stability, estimated propensity scores were clipped at pre-specified bounds before weighting.

For each financial-status indicator, the treatment-specific association was reported as the weighted difference in dropout probability between treated and untreated students. Bootstrap confidence intervals were computed using 120 resamples. These intervals quantify estimation uncertainty under the specified observational design rather than causal certainty.

Two additional estimators were used as robustness checks: counterfactual prediction and augmented inverse probability weighting. These checks assessed whether the estimated directions were stable beyond the main weighting specification. The same estimation pipeline was applied to all data regimes, so differences across regimes reflect differences in the data rather than changes in the analysis procedure.

### 3.3. Causal robustness and sensitivity analysis

The original training set served as the reference regime. For each dataset, we compared the three estimated financial-status associations with the corresponding original-data estimates. Robustness was summarized using direction preservation, signed-order preservation, mean absolute ATE deviation, and the maximum treatment-specific gap. The audit was interpreted sequentially: regimes first had to preserve the original protective or harmful directions; only then were absolute effect sizes compared with the original benchmark. Direction and signed ordering capture qualitative agreement, whereas mean and maximum deviations capture effect-size agreement.

CaP-Eval summarizes causal fidelity with four pre-specified quantities. Let tau_0,k denote the original-training estimate for treatment k and tau_r,k denote the corresponding estimate in non-original regime r, for k = 1,...,K financial-status indicators.

***Definition 1*** (Causal Direction Preservation, CDP). For a given regime, CDP is the fraction of treatment estimates whose sign matches the corresponding original-training estimate. CDP = 1 means all treatment directions are preserved; CDP = 0 means none are preserved.

***Definition 2*** (Causal Rank Fidelity, CRF). CRF is the Spearman rank correlation between the vector of original treatment estimates and the vector of regime-specific treatment estimates. It evaluates whether the relative priority ordering of support-relevant financial indicators is preserved.

***Definition 3*** (Effect-size fidelity). ATE MAD is the mean absolute difference between the original and regime-specific treatment estimates across financial-status indicators. MaxGap is the largest absolute treatment-specific difference from the original benchmark. These metrics quantify agreement with the benchmark; they do not establish the original estimates as ground-truth causal effects.

Three additional checks were conducted. First, the analysis was repeated using Double Machine Learning (DML) with random-forest nuisance models and five-fold cross-fitting. Second, for tuition-fee payment status and debtor status, the IPW analysis was repeated under an alternative adjustment specification that additionally controlled for scholarship-holder status. Third, positivity and weighting behavior were reviewed through treatment prevalence, trimming fractions, standardized mean differences, effective sample sizes, and estimator agreement. Extreme propensity scores were clipped before stabilized weighting. These evaluations were used to verify that the cross-regime comparisons were not driven only by overlap failures or unstable weights.

Because generated data can alter treatment prevalence and covariate support, overlap was treated as a generated-data validity condition rather than as a routine preprocessing step. For each treatment and data regime, the common-support and weighting audit reviewed propensity-score distributions, clipping or trimming fractions, stabilized-weight behavior, effective sample sizes, and post-weighting standardized mean differences. These evaluations were used to flag regimes in which treatment-effect estimates might be numerically stable but not fit for decision-facing use because the generated treatment-covariate structure no longer resembled the original benchmark. Full overlap and weighting summaries are reported in the supplementary materials.

### 3.4. Privacy-risk evaluation of synthetic and distilled datasets

Privacy risk was evaluated as an auxiliary governance screen rather than as a full privacy audit. In institutional data sharing, release decisions often depend on whether generated records remain locally close to identifiable training cases or expose quasi-identifying attribute combinations, even when a generator has a formal privacy-oriented design[49]. We therefore checked exact record reproduction, distance to nearest training records, train-holdout distance dominance, a distance-based membership-inference proxy[50],[51] quasi-identifier exposure, and attribute-disclosure advantage. For the DPGNet regimes, epsilon was treated as a formal privacy-orientation signal tied to the generation mechanism, whereas nearest-record, membership-proxy, quasi-identifier, and attribute-disclosure evaluations were treated as empirical release-governance evidence. These two forms of evidence are complementary but not interchangeable: empirical evaluations do not validate, calibrate, or replace the mathematical privacy budget, and a formal privacy orientation does not guarantee preservation of treatment effects or decision-relevant structure.

## 4. Results

### 4.1. Predictive utility of synthetic and distilled student data

We first evaluated whether each data condition retained sufficient information for dropout prediction on the original held-out test set using the curve-based utility and calibration metrics reported in **Table 1** and **Figure 3**. The original data retained the highest ROC AUC (0.819) and average precision (0.739), followed by distilled data (ROC AUC = 0.802; average precision = 0.709) and TabularGNet (ROC AUC = 0.799; average precision = 0.692). Gaussian Copula and DPGNet epsilon = 1 showed lower but still usable discrimination, with ROC AUC values of 0.755 and 0.751, respectively. Calibration also varied across regimes: Brier scores were lowest for the original data (0.166), TabularGNet (0.173), and DPGNet epsilon = 1 (0.177), and highest for Gaussian Copula (0.208).

The predictive screen shows that DPGNet retained usable discrimination even under the epsilon = 1 condition included in Figure 3. This is important because a privacy-oriented generator should not be rejected or accepted on classifier utility alone. The epsilon grid is evaluated directly in the causal-fidelity sections, where the question is whether changing the privacy budget preserves the financial-status evidence used for support decisions.

Table 1 reports the predictive and calibration conditions visualized in **Figure 3**. The compact predictive figure includes DPGNet epsilon = 1 as the representative privacy-oriented utility screen. Sections 4.3-4.4 then expand DPGNet to epsilon = 1, 5, and 10 because privacy-budget sensitivity is most consequential for treatment-effect direction, magnitude, overlap, and release-governance interpretation.

| Dataset | Evaluation source | ROC AUC | Average precision | Brier score |
|---|---|---|---|---|
| Original | Held-out original test set | 0.819 | 0.739 | 0.166 |

| Dataset | Evaluation source | ROC AUC | Average precision | Brier score |
|---|---|---|---|---|
| Distilled | Held-out original test set | 0.802 | 0.709 | 0.181 |
| TabularGNet | Held-out original test set | 0.799 | 0.692 | 0.173 |
| Gaussian Copula | Held-out original test set | 0.755 | 0.561 | 0.208 |
| DPGNet (epsilon = 1) | Held-out original test set | 0.751 | 0.648 | 0.177 |

**Table 1.** Predictive and calibration utility of original, synthetic, distilled, and DPGNet data on the held-out original test set. Values summarize the curve-based evaluations shown in **Figure 3**.

**Figure 3** complements Table 1 with visual and curve-based utility evaluations. The low-dimensional projections suggest that distilled and TabularGNet data retained more recognizable structure relative to the original data, whereas Gaussian Copula and DPGNet showed stronger distributional reshaping. The precision-recall, ROC, and calibration panels show that DPGNet retained usable predictive signal despite this reshaping, reinforcing that predictive utility and causal-fidelity preservation must be evaluated separately.

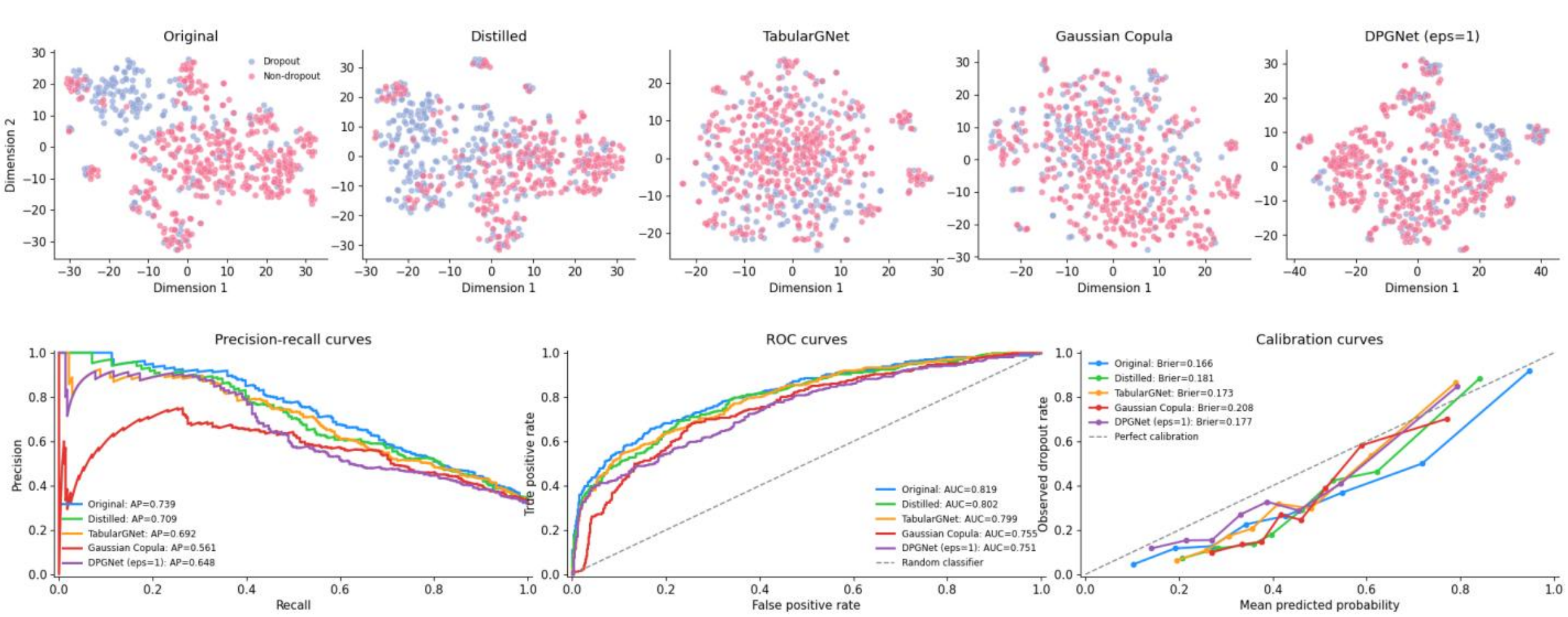

**Figure 3.** Distributional and predictive-utility evaluations for the generated and distilled datasets. The panels compare low-dimensional structure and representative precision-recall, ROC, and calibration behaviour; they are not used as evidence of treatment-effect preservation.

These predictive-utility results motivate the causal-robustness analysis that follows. They show whether each regime remains useful for modelling dropout risk, but they do not establish whether the same regime preserves the financial-status evidence needed for support prioritisation. The next sections therefore evaluate treatment-design validity, marginal representation, treatment-effect preservation, estimator agreement, and privacy exposure under a common observational specification.

## 4.2. Treatment-design validation and marginal representation

We next checked whether the fixed decision-support design remained coherent across the three financial-status indicators. For all treatments, dropout was the outcome, enrollment-time demographic, academic-path, parental-background, and macroeconomic variables formed the baseline adjustment set, and first- and second-semester curricular-unit variables were excluded as post-enrollment descendants. For tuition-fee payment and debtor status, the sensitivity analysis additionally adjusted for scholarship-holder status. The design information is therefore reported in the methods and supplementary sensitivity materials, while the main results focus on whether each non-original regime preserves the treatment-effect evidence needed for dropout-support decisions.

The marginal validation revealed a clear separation between treatment-prevalence agreement and treatment-effect preservation. Gaussian Copula synthesis matched some original treatment rates closely, yet still attenuated the estimated effects. The DPGNet regimes retained the original marginal treatment rates for the three financial-status indicators, but they required larger trimming or overlap adjustments than the non-private baselines, especially for scholarship-holder and debtor status. This pattern shows why privacy-oriented generation should be evaluated on the estimand and the weighting design itself, not only on marginal similarity.

### 4.3. Main treatment effects across datasets

We then estimated the associations between dropout and the three financial-status indicators selected for educational decision support: scholarship-holder status, tuition-fee payment status, and debtor status. The same treatment definitions, baseline covariates, and estimators were used across all regimes. Differences in estimates therefore reflect differences introduced by the data representation rather than changes in the analytic specification.

Across the main IPW estimates and cross-estimator checks, all non-original regimes preserved the original signs of the three financial-status associations, but they differed in effect-size fidelity. Supplementary Table S1 reports treatment prevalence, unadjusted treated-minus-untreated dropout differences, stabilized IPW estimates, and trimming fractions across all regimes. The original benchmark remained strongly protective for scholarship-holder status (ATE = -0.194) and tuition-fee payment (ATE = -0.572), and harmful for debtor status (ATE = 0.343). Distilled and TabularGNet regimes preserved all three signs but attenuated the larger tuition-payment and debtor contrasts; Gaussian Copula preserved signs with stronger attenuation; and DPGNet preserved all three signs across epsilon = 1, 5, and 10 while showing epsilon-dependent magnitude changes.

Table 2 reports the stabilized IPW estimates under the fixed observational specification and the corresponding magnitude-change comparison. The distilled regime remained close to the original benchmark for scholarship-holder status but attenuated the larger tuition-payment and debtor contrasts. TabularGNet preserved all three directions while compressing the original effect scale, and Gaussian Copula showed stronger attenuation. DPGNet showed a distinct privacy-oriented profile: all epsilon levels preserved direction and rank, epsilon = 10 had the smallest aggregate IPW deviation, and epsilon = 1 and epsilon = 5 amplified several financial-status contrasts. In decision support, direction is only the first screen; inflated effect sizes can over-prioritize some financial-status groups and misallocate scarce advising or financial-aid resources.

| Treatment | Dataset | Stabilized IPW ATE | 95% Bootstrap CI | ATE magnitude change (%) |
|---|---|---|---|---|
| Scholarship holder | Original | -0.194 | [-0.230, -0.165] | +0.0 |
| | Distilled | -0.212 | [-0.295, -0.139] | +9.3 |
| | TabularGNet | -0.090 | [-0.127, -0.049] | -53.6 |

| Treatment | Dataset | Stabilized IPW ATE | 95% Bootstrap CI | ATE magnitude change (%) |
|---|---|---|---|---|
| | Gaussian Copula | -0.064 | [-0.096, -0.030] | -67.0 |
| | DPGNet (epsilon = 1) | -0.271 | [-0.321, -0.214] | +39.7 |
| | DPGNet (epsilon = 5) | -0.321 | [-0.384, -0.262] | +65.5 |
| | DPGNet (epsilon = 10) | -0.090 | [-0.144, 0.008] | -53.6 |
| Tuition fees up to date | Original | -0.572 | [-0.605, -0.531] | +0.0 |
| | Distilled | -0.374 | [-0.423, -0.303] | -34.6 |
| | TabularGNet | -0.304 | [-0.337, -0.262] | -46.9 |
| | Gaussian Copula | -0.134 | [-0.192, -0.092] | -76.6 |
| | DPGNet (epsilon = 1) | -0.678 | [-0.717, -0.633] | +18.5 |
| | DPGNet (epsilon = 5) | -0.706 | [-0.738, -0.641] | +23.4 |
| | DPGNet (epsilon = 10) | -0.480 | [-0.548, -0.411] | -16.1 |
| Debtor | Original | 0.343 | [0.286, 0.374] | +0.0 |
| | Distilled | 0.194 | [0.093, 0.292] | -43.4 |
| | TabularGNet | 0.140 | [0.106, 0.181] | -59.2 |
| | Gaussian Copula | 0.076 | [0.027, 0.122] | -77.8 |
| | DPGNet (epsilon = 1) | 0.543 | [0.487, 0.625] | +58.3 |
| | DPGNet (epsilon = 5) | 0.391 | [0.309, 0.477] | +14.0 |
| | DPGNet (epsilon = 10) | 0.346 | [0.281, 0.408] | +0.9 |

**Table 2.** Stabilized IPW treatment-effect estimates and ATE magnitude change across datasets. ATE magnitude change was calculated relative to the original-data estimate within each treatment as (|ATE_regime|/|ATE_original| - 1) × 100. Positive values indicate amplification relative to the original benchmark, whereas negative values indicate attenuation; sign reversals are interpreted separately in the text.

**Figure 4** re-expresses the main IPW treatment-effect results as signed deviations from the original benchmark. The figure follows the two-stage audit rule used throughout CaP-Eval. First, each non-original regime is assessed for directional validity: scholarship-holder status and up-to-date tuition payment should remain protective, and debtor status should remain harmful. Second, among regimes that pass this sign screen, deviation from zero shows how far each generated or distilled estimate moves away from the original observational benchmark.

This deviation view makes amplification and attenuation more visible than raw ATE curves. Distilled data remained close for scholarship-holder status but attenuated tuition-payment and debtor effects. TabularGNet and Gaussian Copula compressed the treatment effects toward zero. DPGNet epsilon = 10 showed the smallest aggregate deviation, driven by near-perfect debtor preservation and moderate tuition-payment attenuation, but it still attenuated scholarship-holder status. DPGNet epsilon = 1 and epsilon = 5 preserved directions while amplifying several protective and harmful contrasts; in practice, this could overstate the urgency of some financial-support signals and divert resources from students with comparable need.

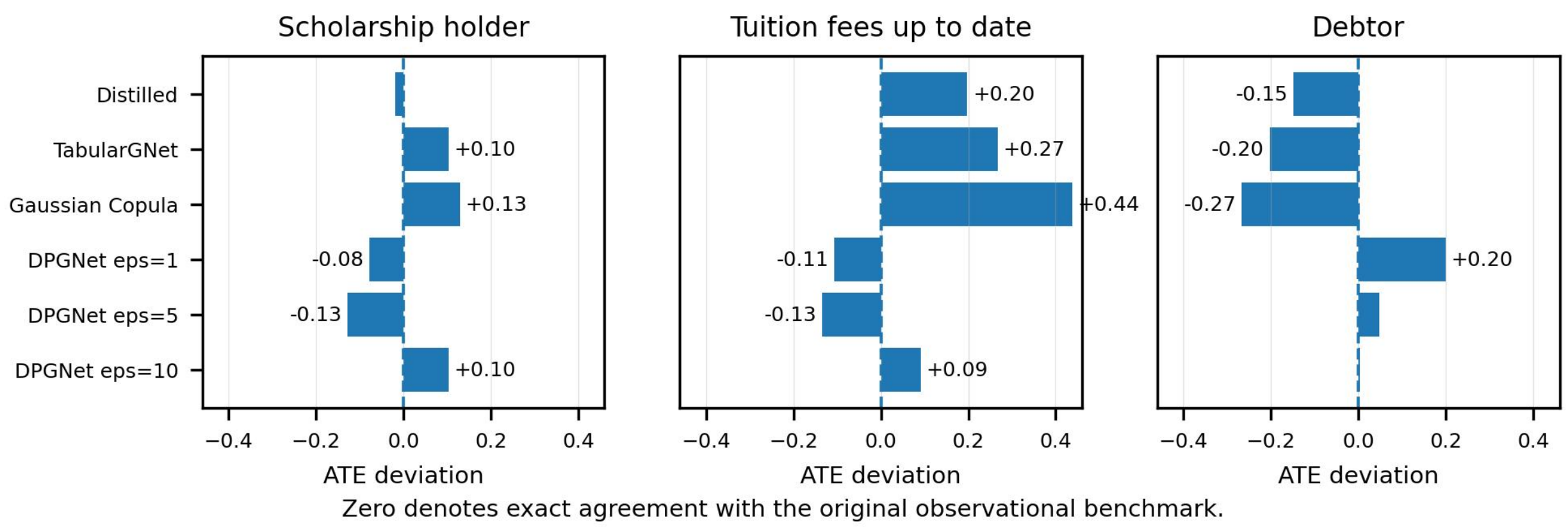

**Figure 4.** Signed treatment-effect deviations from the original IPW benchmark. Values are ATE_regime - ATE_original; zero means exact agreement. Direction is the entry screen; among direction-preserving regimes, distance from zero shows treatment-specific attenuation or amplification relevant to support prioritization.

### 4.4. Aggregate robustness and sensitivity validation

After estimating financial-status associations for each indicator, we assessed aggregate causal fidelity. Table 3 summarizes direction preservation, signed-order preservation, mean absolute ATE deviation, and the largest treatment-specific gap from the original estimate. Because all non-original regimes passed the directional screen and preserved signed rank ordering, the decisive comparison shifted to effect-size fidelity. DPGNet epsilon = 10 showed the strongest aggregate IPW fidelity (ATE MAD = 0.067; maximum gap = 0.104), followed by DPGNet epsilon = 5, distilled data, DPGNet epsilon = 1, TabularGNet, and Gaussian Copula.

| Regime | ATE MAD | Max abs gap | Spearman rho | Direction preservation |
|---|---|---|---|---|
| Distilled | 0.122 | 0.199 | 1.0 | 1.000 |
| TabularGNet | 0.192 | 0.269 | 1.0 | 1.000 |
| Gaussian Copula | 0.278 | 0.439 | 1.0 | 1.000 |
| DPGNet (epsilon = 1) | 0.127 | 0.200 | 1.0 | 1.000 |
| DPGNet (epsilon = 5) | 0.102 | 0.133 | 1.0 | 1.000 |
| DPGNet (epsilon = 10) | 0.067 | 0.104 | 1.0 | 1.000 |

**Table 3.** Overall causal fidelity across non-original data regimes. MAD and maximum absolute gap were calculated against the original-data IPW benchmark across the three financial-status indicators.

Effect-size fidelity separated regimes that otherwise agreed in direction. DPGNet epsilon = 10 was closest to the original IPW benchmark (ATE MAD = 0.067; maximum gap = 0.104), but still attenuated scholarship-holder status. DPGNet epsilon = 5 had strong aggregate fidelity (ATE MAD = 0.102; maximum gap = 0.133) while amplifying scholarship-holder and tuition-payment contrasts. Distilled data remained competitive, whereas TabularGNet and Gaussian Copula showed larger deviations. Qualitative agreement therefore did not imply reliable effect-size preservation.

We next evaluated sensitivity to an alternative adjustment specification, because propensity-score estimates can depend on covariate choice and weighting specification[14]. This check was applied to up-to-date tuition status and debtor status because these variables are more closely connected to other financial-status conditions than scholarship-holder status. For up-to-date tuition status, the original IPW estimate changed from -0.572 to -0.551, an absolute change of 0.021. The corresponding changes were 0.008 for distilled data, 0.010 for TabularGNet, 0.001 for Gaussian Copula, 0.002 for DPGNet epsilon = 1, 0.005 for DPGNet epsilon = 5, and 0.001 for DPGNet epsilon = 10. For debtor status, the original estimate changed from 0.343 to 0.325; the largest non-original change was observed for DPGNet epsilon = 1 (0.051), while all directions remained stable.

The aggregate comparison was also repeated using Double Machine Learning with nuisance-model residualization and cross-fitting. The DML-based fidelity results supported the same broad interpretation: DPGNet epsilon = 10 was closest to the original estimates (DML MAD = 0.039), followed closely by distilled data (0.043), TabularGNet (0.090), DPGNet epsilon = 5 (0.117), DPGNet epsilon = 1 (0.138), and Gaussian Copula (0.151). DML reduced some absolute deviations relative to IPW, but it did not change the main conclusion that generated data must be checked on treatment-effect fidelity rather than assumed valid from predictive utility or privacy orientation alone.

Supplementary overlap and weighting evaluations after stabilized IPW showed acceptable post-weighting balance across regimes, with low mean and maximum standardized mean differences after weighting. However, generated-data overlap still mattered: the DPGNet regimes required non-zero trimming and had lower effective sample sizes in several treatment conditions, especially for scholarship-holder and debtor status at tighter or intermediate privacy budgets. Acceptable balance therefore remains necessary but not sufficient; it must be interpreted alongside causal-fidelity, magnitude, and privacy-governance evidence.

Overall, the supplementary tables support the same interpretation. DPGNet, distilled data, TabularGNet, and Gaussian Copula all preserved the direction and signed ordering of the financial-status effects. The main difference was effect-size fidelity: DPGNet epsilon = 10 and distilled data were strongest overall, DPGNet epsilon = 1 and epsilon = 5 showed greater magnitude amplification for some treatments, TabularGNet attenuated the contrasts, and Gaussian Copula showed the strongest compression. The conclusion was stable to the alternative adjustment specification and the DML check.

### 4.5. Privacy-review and sensitivity evaluations

The privacy component is used as a release-governance screen rather than as a full attack-based privacy audit. The analysis therefore focuses on empirical disclosure indicators directly relevant to institutional data sharing. The goal is to identify whether regimes that preserve decision-relevant causal structure also retain local training-record proximity signal, and whether more privacy-oriented regimes remain usable for causal decision support. Figure 5 focuses on distance-based privacy evaluations; exact-match, attribute-disclosure, and membership-proxy checks are summarized in the supplementary materials because they support the same governance interpretation without changing the main empirical conclusion.

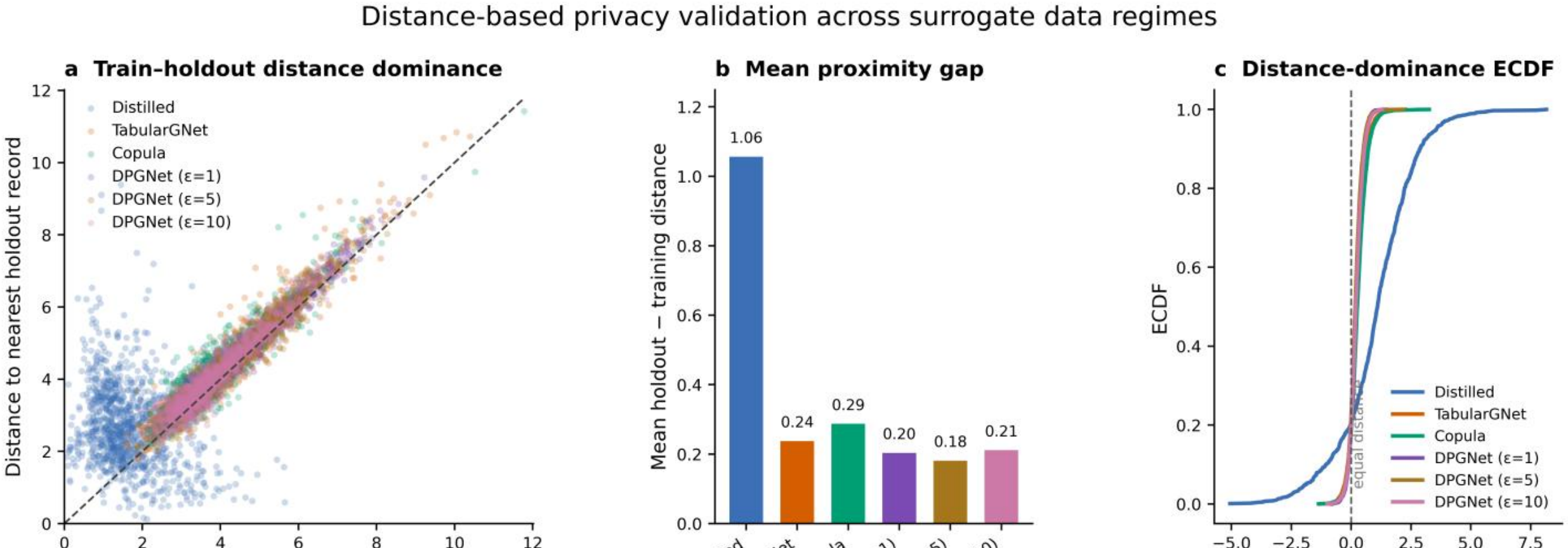

**Figure 5**. Distance-based privacy evaluations. The panels compare each generated record's nearest-training and nearest-holdout distances across data regimes.

The three distance views answer complementary questions. The dominance scatter asks whether generated records are systematically closer to training records than to holdout records. The mean proximity gap summarizes that tendency at the regime level, and the ECDF shows whether the pattern reflects a few local cases or a broader distributional shift. Read together, these panels show that the distilled data retain the strongest training-proximal structure, whereas DPGNet regimes move farther from training records while still preserving the decision-relevant treatment directions in the causal audit.

The broader empirical evaluations reinforce the same governance interpretation. Exact-match checks and average distance-based membership proxies were broadly reassuring for non-original regimes, but distilled data remained the most locally training-proximal among the non-original datasets. DPGNet regimes provided the privacy-oriented comparison condition and showed lower local proximity than distilled data, yet they did not eliminate the need for release review because trimming, overlap, and effect-size behavior varied across epsilon levels. Formal privacy orientation, empirical disclosure screening, and causal fidelity should therefore be reported separately rather than collapsed into a single claim that one regime is simply most private or most valid.

This evidence supports a layered governance interpretation. High-fidelity distilled data are defensible for controlled internal validation, estimator stress-testing, and restricted institutional research, but their stronger local training-record proximity signal argues against broad external release without additional controls. Statistical or adversarial synthetic data are more appropriate for software demonstrations, teaching, and workflow testing where exact treatment-effect magnitudes are not acted on. Under this implementation, DPGNet epsilon = 10 may be useful for privacy-oriented workflow testing and controlled internal causal prototyping, but it should not be used for student-facing causal decisions unless the relevant treatment structure is revalidated against original institutional data.

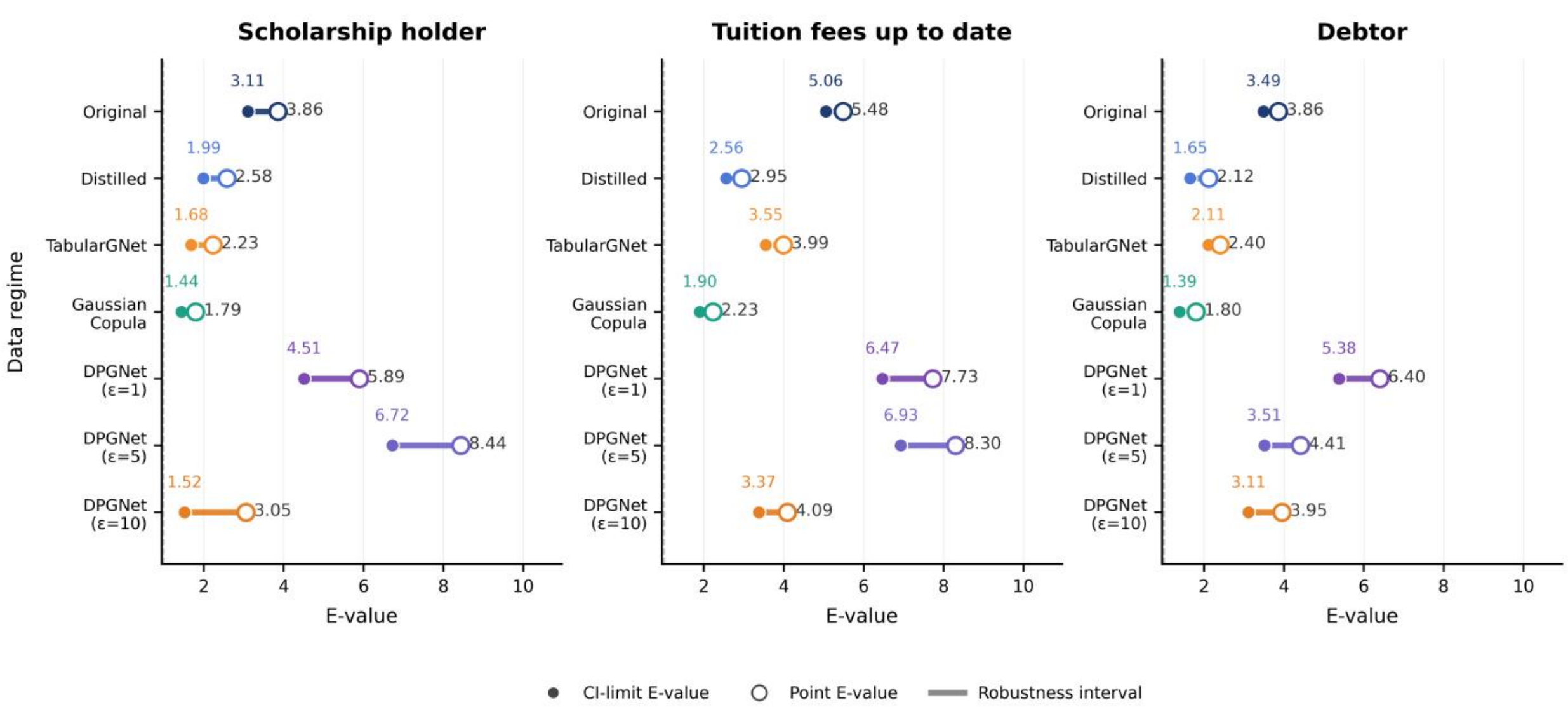

**Figure 6**. Sensitivity to unmeasured confounding based on E-value evaluations[52]. Filled points show confidence-limit E-values and open points show point-estimate E-values; larger values indicate that stronger unmeasured confounding would be required to explain away the observed association under the specified observational design.

The E-value evaluations support the treatment-effect interpretation only as a sensitivity analysis for unmeasured confounding. The original, distilled, and DPGNet estimates for tuition-fee payment and debtor status show stronger robustness to unmeasured confounding than the more attenuated Gaussian Copula estimates. E-values are therefore used as a sensitivity screen for the observational benchmark, not as proof that the original estimates are unbiased causal effects.

### 4.6. Additional robustness checks

To keep the main results focused, the remaining robustness analyses are summarized here and reported in the supplementary tables. The Double Machine Learning (DML) analysis, using random-forest nuisance models and five-fold cross-fitting, confirmed the same broad ordering as the IPW analysis: DPGNet epsilon = 10 and distilled data were closest to the original estimates (DML MAD = 0.039 and 0.043, respectively), followed by TabularGNet (0.090), DPGNet epsilon = 5 (0.117), DPGNet epsilon = 1 (0.138), and Gaussian Copula (0.151). All DPGNet regimes preserved the original treatment-effect directions under DML.

Support-size ablation. The support-size ablation of distilled data also supported the main interpretation. Even the smallest distilled support set (n = 10) preserved all three treatment-effect directions and achieved an ATE MAD of 0.221, outperforming the full Gaussian Copula regime (ATE MAD = 0.278). Fidelity improved with support size and stabilized around 300-1,000 support points; the 750- and 1,000-point support sets achieved ATE MAD values of 0.118 and 0.122, respectively. This suggests that a 300-500 point support set may be a practical compromise for controlled internal validation, although compact supports still require disclosure review and access controls. Detailed ablation results are provided in Supplementary Table S4.

Subgroup consistency. Exploratory subgroup checks across gender, age group, and international student status followed the aggregate interpretation: distilled data generally showed strong subgroup fidelity, TabularGNet and Gaussian Copula showed larger attenuation, and DPGNet preserved subgroup sign consistency while showing epsilon-dependent variation in subgroup rank fidelity and CATE magnitude. These results are treated as supporting evidence only because some treatment-subgroup cells were small and no multiplicity-adjusted confirmatory inference was claimed.

## 5. Discussion

The findings sharpen synthetic-data evaluation in learning analytics by adding decision-facing causal fidelity to utility and privacy screening. Classifier performance, distributional resemblance, local proximity, privacy orientation, and treatment-effect magnitude diverged in ways that matter for institutional decisions. The results therefore reveal a non-monotonic governance profile rather than a simple utility-privacy frontier: distilled data were faithful but training-proximal, whereas the evaluated DPGNet configuration showed its strongest causal-fidelity profile at epsilon = 10 and exposed amplification risks at epsilon = 1 and epsilon = 5. The support-size analysis adds a practical dimension by showing when compact distilled supports can preserve treatment ordering while still requiring disclosure review.

The main lesson is practical rather than methodological for its own sake. Universities do not need synthetic student data only to reproduce a classifier score; they need it to support decisions made under budget, staffing, and privacy constraints. A financial-aid office may ask whether payment-status changes should trigger outreach. An advising unit may ask whether debtor status identifies a group needing coordinated academic and financial support. Those questions depend on the direction and scale of financial-status associations, not only on whether a model ranks dropouts above non-dropouts.

**Figure 4** carries much of the paper's substantive argument because it separates the entry condition from the operational question. The entry condition is directional validity: generated data should preserve whether each financial-status signal is protective or harmful. The operational question is magnitude: whether the estimated effect is close enough to guide support intensity. All non-original regimes passed the first screen, but their magnitude profiles differed sharply.

The practical risk is effect-size distortion. Attenuated estimates could understate the expected benefit of payment-plan outreach or under-prioritize financially vulnerable students. Amplified estimates create the opposite risk: institutions may over-prioritize a group or overstate the urgency of a support signal. For DPGNet, epsilon = 1 and epsilon = 5 amplified several contrasts by approximately 18%-66%, whereas epsilon = 10 was closest overall but still attenuated scholarship-holder status. Even the strongest privacy-oriented candidate therefore requires treatment-specific magnitude and overlap review before operational use.

For decision-support systems, treatment-effect distortion carries an organizational cost. Sign inversion remains the most severe failure mode because it can become a form of algorithmic policy inversion. Magnitude distortion is subtler but still consequential: it can change outreach intensity, advising priority, or the perceived urgency of payment-plan support. Repeated mismatches between student need and data-supported intervention may erode advisor reliance, student perceptions of fairness, and confidence in institutional decision support[53],[54].

**Figure 5** shows the privacy side of the same conclusion. The distilled set remained highly faithful to the original treatment-effect pattern and also retained the strongest local training-record proximity signal in the empirical privacy evaluations. This does not prove membership disclosure, and it does not mean distillation is unsafe in every setting. It means that the mechanism producing strong analytic fidelity may also preserve fine-grained training structure, which should trigger disclosure review before external sharing. DPGNet showed a different profile, combining full direction preservation with lower local proximity but epsilon-dependent effect-size behavior.

For education data, that point is not merely technical. A small group of students can be distinctive because of the joint pattern of course, age, parental background, scholarship status, debt, and academic pathway. A record that is not an exact copy can still sit close enough to a real training case to deserve scrutiny, especially when data are shared outside the institution. Together, the privacy evaluations support a tiered-use rule: high-fidelity distilled data are defensible for controlled internal work, whereas broader release requires stronger privacy evidence than average-distance summaries or exact-duplicate checks.

A plausible mechanism for the stronger causal fidelity of distillation is that the kernel-based procedure optimizes support points against the prediction geometry of the full training set, thereby retaining aspects of the conditional structure that underlie both predictive and treatment-outcome signals. In contrast, adversarial and copula synthesis primarily aim to reproduce marginal or joint distributional structure, and privacy-oriented generation can alter conditional associations through privacy-budgeted perturbation and mixed-type post-processing. DPGNet epsilon = 10 may have preserved the relevant conditional structure better than the tighter privacy budgets in this implementation, but epsilon = 1 and epsilon = 5 also amplified some treatment contrasts. This interpretation is suggestive rather than definitive: the study does not prove that distillation or DPGNet generally preserves causal structure, and the local-structure retention that helps fidelity is also why generated data require disclosure review.

This mechanism also clarifies why the timing-aware design is used as an audit standard rather than as a constraint imposed on the generators. The distillation, adversarial, copula, and DPGNet procedures treat the tabular data largely as a flat matrix; they are not instructed to preserve an enrollment-to-semester temporal ordering, enforce causal parent-child relations, or protect the financial-status estimand. Any sign reversal, effect attenuation, or effect amplification should therefore be read as evidence that a flat generation process may have altered decision-relevant conditional structure, not as proof of a single underlying failure mechanism. CaP-Eval does not assume that generated data are causally aware. It asks a downstream decision-validity question: after flat generation, does the resulting dataset still support the same timing-aware evidence that would be used in the original benchmark?

The education setting makes the causal-fidelity/local-proximity tension especially concrete because the variables of interest are tied to real support actions--financial-aid decisions, payment-plan outreach, and debt advising--not only to benchmark accuracy. Distillation can reduce direct exposure of raw examples, but critiques of "privacy for free" show that local evaluation can change the conclusion[16],[17]. Studies of synthetic-data training and distillation similarly warn that average-case privacy metrics may miss vulnerable records or transferred memorized information[18],[40]. The present results therefore support a cautious interpretation: high analytic fidelity is useful, but it should trigger stronger disclosure review rather than an assumption of safe release.

Practically, CaP-Eval should be read as a release-and-use screen rather than as a single pass-or-fail label. Distilled data meet a strong decision-validity screen but require restricted access because they also retain the strongest training-proximal signal. In this study, DPGNet epsilon = 10 is the strongest privacy-oriented candidate for controlled causal prototyping, but it still requires overlap, magnitude, implementation, and privacy review before operational use. TabularGNet and Gaussian Copula are more appropriate for dashboards, analyst training, software testing, and workflow rehearsal where exact treatment-effect magnitudes are not acted on. No non-original regime should be used for student-facing causal decision support without revalidation against original institutional data.

Beyond dropout support, the same CaP-Eval logic can be extended to other decision-facing learning-analytics settings, such as course recommendation, advising-resource allocation, scholarship prioritization, and early-alert workflow testing, wherever generated data are used to support institutional decisions rather than only model development[28],[49].

Internal validity. The analysis is observational and depends on the no-unmeasured-confounding assumption conditional on measured baseline covariates. Scholarship-holder status has the clearest interpretation under this design; tuition-fee payment status and debtor status may partly reflect institutional policies or student circumstances that are not fully captured by the available covariates. The timing-aware DAG, overlap/weight evaluations, and E-value sensitivity checks mitigate but do not eliminate this concern. The subgroup CATE analysis should also be interpreted as exploratory: some subgroup-treatment cells were small, especially for debtor status, and no multiplicity-adjusted confirmatory subgroup inference is claimed.

**External validity.** The study uses a single formal higher-education dataset from one national context and one evaluated DPGNet implementation. The dataset contains several common features of administrative student data--mixed variable types, class imbalance, financial-status confounding, and post-treatment academic measures--but the ordering of data regimes should not be treated as universal. Future multi-institutional replication should test whether the observed fidelity ordering holds across different national contexts, student populations, financial-aid systems, administrative data infrastructures, and differentially private generator implementations.

**Privacy and generator scope.** The DPGNet results should be interpreted as an implementation-specific privacy-oriented stress test rather than as an optimized benchmark for all differentially private synthetic-data generators. The implementation did not use teacher aggregation, so teacher partitioning and teacher counts were not part of this regime. A central transparency limitation is that optimizer-level privacy details--including clipping norms, noise calibration schedules, accountant internals, and architecture-specific training settings--were not exposed and could not be audited directly from generated outputs. Future releases should report these quantities explicitly and combine formal privacy accounting with stronger attack models. The empirical privacy evaluations in this study are governance screens, not a complete privacy audit, and they should not be interpreted as substitutes for formal privacy guarantees.

For practice, the safest conclusion is modest. Generated student data can be helpful, but their acceptable use depends on the decision being made. Teaching examples, code development, and pipeline testing require a different evidentiary standard from budget allocation, automated outreach, or eligibility rules. Before non-original data are used for student-facing decisions, institutions should check whether the relevant support signals survive the generation process and whether the privacy review matches the release context.

## 6. Conclusion

This study introduced CaP-Eval, a decision-facing audit workflow for evaluating whether synthetic and distilled student data preserve the evidence needed for dropout decision support. Across seven data conditions and four estimators, DPGNet epsilon = 10 and distilled data most closely preserved the original financial-status treatment-effect structure, but with different governance profiles: DPGNet showed lower local training-record proximity and epsilon-dependent magnitude behavior, whereas distilled data retained stronger local training-record proximity. TabularGNet and Gaussian Copula preserved directions with greater attenuation. The central conclusion is not that one generator is universally best. Rather, DPGNet epsilon = 10 was the strongest privacy-oriented candidate in this study, but it still attenuated scholarship-holder status and depends on implementation transparency, overlap, local proximity, and institutional decision context before release or use.

For the learning analytics, educational data mining, and applied expert-systems communities, the practical implication is that use-specific causal validation should become standard practice before generated student data are used for institutional decisions. CaP-Eval can serve as a reusable template for that validation: define the decision-relevant treatments, apply a timing-aware adjustment design, test direction and rank fidelity across estimators, audit overlap and weighting behavior, and interpret privacy evaluations as governance evidence rather than as automatic permission for release[10]. For institutional researchers and data protection officers, the tiered-use message is deliberately conservative: high-fidelity distilled data are most defensible for controlled internal validation, DPGNet epsilon = 10 is the most promising privacy-oriented option for controlled causal prototyping after revalidation, statistical or adversarial synthetic data are more appropriate for demonstrations and workflow testing, and no non-original data should be used for student-facing causal decisions unless the relevant treatment structure is revalidated.

**Declaration of competing interests**

The authors declare that they have no known competing financial interests or personal relationships that could have appeared to influence the work reported in this paper.

**Data availability**

The Students' Dropout and Academic Success dataset is publicly available from the UCI Machine Learning Repository. The analysis code and evaluation protocols will be deposited in a public GitHub or OSF repository upon acceptance.

**Code availability**

The analysis code used in this study will be made available upon request.

## Supplementary Materials

The supplementary materials contain tabular robustness checks used to support the main-text results. They do not include additional supplementary figures in this manuscript file.

**Table S1** reports treatment prevalence, unadjusted contrasts, stabilized IPW estimates, bootstrap intervals, and trimming fractions across all data regimes.

| Treatment | Dataset | Treatment rate | Naive diff. | Stab. IPW ATE | 95% Bootstrap CI | Trim |
|---|---|---|---|---|---|---|
| Scholarship holder | Original | 0.245 | -0.265 | -0.194 | [-0.230, -0.165] | 0 |
| | Distilled | 0.214 | -0.271 | -0.212 | [-0.295, -0.139] | 0 |
| | TabularGNet | 0.162 | -0.110 | -0.090 | [-0.127, -0.049] | 0 |
| | Gaussian Copula | 0.248 | -0.091 | -0.064 | [-0.096, -0.030] | 0 |
| | DPGNet (epsilon = 1) | 0.245 | -0.265 | -0.271 | [-0.321, -0.214] | 0.220 |
| | DPGNet (epsilon = 5) | 0.245 | -0.265 | -0.321 | [-0.384, -0.262] | 0.412 |
| | DPGNet (epsilon = 10) | 0.245 | -0.265 | -0.090 | [-0.144, 0.008] | 0.198 |
| Tuition fees up to date | Original | 0.883 | -0.618 | -0.572 | [-0.605, -0.531] | 0 |
| | Distilled | 0.826 | -0.414 | -0.374 | [-0.423, -0.303] | 0 |
| | TabularGNet | 0.798 | -0.347 | -0.304 | [-0.337, -0.262] | 0 |
| | Gaussian Copula | 0.881 | -0.142 | -0.134 | [-0.192, -0.092] | 0 |
| | DPGNet (epsilon = 1) | 0.883 | -0.618 | -0.678 | [-0.717, -0.633] | 0.064 |
| | DPGNet (epsilon = 5) | 0.883 | -0.618 | -0.706 | [-0.738, -0.641] | 0.349 |
| | DPGNet (epsilon = 10) | 0.883 | -0.618 | -0.480 | [-0.548, -0.411] | 0.102 |
| Debtor | Original | 0.117 | 0.349 | 0.343 | [0.286, 0.374] | 0 |
| | Distilled | 0.154 | 0.211 | 0.194 | [0.093, 0.292] | 0 |
| | TabularGNet | 0.180 | 0.160 | 0.140 | [0.106, 0.181] | 0 |
| | Gaussian Copula | 0.121 | 0.079 | 0.076 | [0.027, 0.122] | 0 |
| | DPGNet (epsilon = 1) | 0.117 | 0.349 | 0.543 | [0.487, 0.625] | 0.369 |
| | DPGNet (epsilon = 5) | 0.117 | 0.349 | 0.391 | [0.309, 0.477] | 0.599 |
| | DPGNet (epsilon = 10) | 0.117 | 0.349 | 0.346 | [0.281, 0.408] | 0.234 |

**Table S1.** Treatment prevalence and IPW estimates

**Table S2** summarizes the alternative-adjustment sensitivity analysis for tuition-fee payment and debtor status.

| Treatment | Dataset | Main IPW ATE | Sensitivity IPW ATE | Abs change |
|---|---|---|---|---|
| Tuition fees up to | Original | -0.572 | -0.551 | 0.021 |

| Treatment | Dataset | Main IPW ATE | Sensitivity IPW ATE | Abs change |
|---|---|---|---|---|
| date | Distilled | -0.374 | -0.366 | 0.008 |
| | TabularGNet | -0.304 | -0.294 | 0.010 |
| | Gaussian Copula | -0.134 | -0.133 | 0.001 |
| | DPGNet (epsilon = 1) | -0.678 | -0.676 | 0.002 |
| | DPGNet (epsilon = 5) | -0.706 | -0.710 | 0.005 |
| | DPGNet (epsilon = 10) | -0.480 | -0.478 | 0.001 |
| Debtor | Original | 0.343 | 0.325 | 0.017 |
| | Distilled | 0.194 | 0.192 | 0.002 |
| | TabularGNet | 0.140 | 0.135 | 0.006 |
| | Gaussian Copula | 0.076 | 0.076 | 0.001 |
| | DPGNet (epsilon = 1) | 0.543 | 0.594 | 0.051 |
| | DPGNet (epsilon = 5) | 0.391 | 0.401 | 0.011 |
| | DPGNet (epsilon = 10) | 0.346 | 0.334 | 0.013 |

**Table S2.** Alternative-adjustment sensitivity estimates

**Table S3** reports the Double Machine Learning fidelity summary used to evaluate whether the causal-fidelity ordering was sensitive to the main IPW specification.

| Regime | DML ATE MAD | DML max abs gap | DML Spearman rho | DML direction preservation |
|---|---|---|---|---|
| Distilled | 0.043 | 0.060 | 1 | 1 |
| TabularGNet | 0.090 | 0.096 | 1 | 1 |
| Gaussian Copula | 0.151 | 0.230 | 1 | 1 |
| DPGNet (epsilon = 1) | 0.138 | 0.190 | 1 | 1 |
| DPGNet (epsilon = 5) | 0.117 | 0.174 | 1 | 1 |
| DPGNet (epsilon = 10) | 0.039 | 0.100 | 1 | 1 |

**Table S3**. Double Machine Learning fidelity summary

**Table S4** reports weighting, overlap, and balance evaluations used to assess whether cross-regime comparisons were affected by unstable propensity weighting.

| Treatment | Dataset | Mean SMD before | Mean SMD after | Max SMD before | Max SMD after | Effective n | Overlap n | Trim fraction |
|---|---|---|---|---|---|---|---|---|
| Scholarship holder | Original | 0.179 | 0.009 | 0.488 | 0.032 | 2604.4 | 3539 | 0 |
| | Distilled | 0.219 | 0.027 | 0.520 | 0.090 | 606.8 | 1000 | 0 |
| | TabularGNet | 0.076 | 0.004 | 0.285 | 0.017 | 3053.6 | 3539 | 0 |
| | Gaussian Copula | 0.101 | 0.003 | 0.238 | 0.009 | 3002.0 | 3539 | 0 |
| | DPGNet (epsilon = 1) | 0.516 | 0.073 | 1.381 | 0.226 | 715.3 | 2762 | 0.220 |
| | DPGNet (epsilon = 5) | 0.452 | 0.063 | 1.437 | 0.247 | 415.7 | 2082 | 0.412 |
| | DPGNet (epsilon = 10) | 0.374 | 0.086 | 1.074 | 0.278 | 440.1 | 2837 | 0.198 |
| Tuition fees up to date | Original | 0.147 | 0.011 | 0.488 | 0.054 | 2613.7 | 3539 | 0 |
| | Distilled | 0.123 | 0.007 | 0.371 | 0.017 | 780.9 | 1000 | 0 |
| | TabularGNet | 0.097 | 0.007 | 0.466 | 0.038 | 2736.5 | 3539 | 0 |
| | Gaussian Copula | 0.058 | 0.002 | 0.156 | 0.006 | 3155.1 | 3539 | 0 |
| | DPGNet (epsilon = 1) | 0.454 | 0.034 | 0.934 | 0.070 | 568.6 | 3311 | 0.064 |

| Treatment | Dataset | Mean SMD before | Mean SMD after | Max SMD before | Max SMD after | Effective n | Overlap n | Trim fraction |
|---|---|---|---|---|---|---|---|---|
| | DPGNet (epsilon = 5) | 0.449 | 0.057 | 1.995 | 0.200 | 374.3 | 2303 | 0.349 |
| | DPGNet (epsilon = 10) | 0.311 | 0.081 | 1.020 | 0.159 | 564.8 | 3179 | 0.102 |
| Debtor | Original | 0.165 | 0.005 | 0.408 | 0.028 | 2777.5 | 3539 | 0 |
| | Distilled | 0.170 | 0.023 | 0.465 | 0.071 | 667.3 | 1000 | 0 |
| | TabularGNet | 0.085 | 0.004 | 0.239 | 0.024 | 2913.9 | 3539 | 0 |
| | Gaussian Copula | 0.064 | 0.002 | 0.159 | 0.006 | 3136.4 | 3539 | 0 |
| | DPGNet (epsilon = 1) | 0.410 | 0.060 | 1.240 | 0.359 | 351.3 | 2233 | 0.369 |
| | DPGNet (epsilon = 5) | 0.590 | 0.110 | 1.572 | 0.290 | 213.6 | 1420 | 0.599 |
| | DPGNet (epsilon = 10) | 0.374 | 0.056 | 1.068 | 0.146 | 372.0 | 2712 | 0.234 |

**Table S4.** Weighting, overlap, and balance evaluations